\documentclass[letterpaper]{article} % DO NOT CHANGE THIS
\usepackage{aaai2026}  % DO NOT CHANGE THIS
\usepackage{times}  % DO NOT CHANGE THIS
\usepackage{helvet}  % DO NOT CHANGE THIS
\usepackage{courier}  % DO NOT CHANGE THIS
\usepackage[hyphens]{url}  % DO NOT CHANGE THIS
\usepackage{amssymb} % 或
\usepackage{amsfonts}
\usepackage{amsmath} 
\usepackage{graphicx} % DO NOT CHANGE THIS
\usepackage{natbib}  % DO NOT CHANGE THIS AND DO NOT ADD ANY OPTIONS TO IT
\usepackage{caption} % DO NOT CHANGE THIS AND DO NOT ADD ANY OPTIONS TO IT
\usepackage{subfigure}
\usepackage{color, colortbl}

\usepackage{algorithm}
\usepackage{algorithmic}
\newtheorem{theorem}{Theorem}

\newtheorem{remark}{Remark}

\newtheorem{proposition}{Proposition}
\usepackage{booktabs}

\usepackage{makecell}
\usepackage{multirow}
\usepackage{newfloat}
\usepackage{listings}
\DeclareCaptionStyle{ruled}{labelfont=normalfont,labelsep=colon,strut=off} % DO NOT CHANGE THIS
\floatstyle{ruled}
\newfloat{listing}{tb}{lst}{}
\floatname{listing}{Listing}
\title{Multi-modal Dynamic Proxy Learning for Personalized Multiple Clustering}
\author{
Jinfeng Xu\textsuperscript{\rm 1},
Zheyu Chen\textsuperscript{\rm 2},
Shuo Yang\textsuperscript{\rm 1},
Jinze Li\textsuperscript{\rm 1},
Ziyue Peng\textsuperscript{\rm 3},
Zewei Liu\textsuperscript{\rm 1},
Hewei Wang\textsuperscript{\rm 4},
Jiayi Zhang\textsuperscript{\rm 5},
Edith C. H. Ngai\textsuperscript{\rm 1}\thanks{Corresponding authors}
}
\affiliations {
\textsuperscript{\rm 1}The University of Hong Kong,\\
\textsuperscript{\rm 2}Beijing Institute of Technology,\\
\textsuperscript{\rm 3}University of Macau,\\
\textsuperscript{\rm 4}Carnegie Mellon University,\\
\textsuperscript{\rm 5}University of Nottingham,\\
\{jinfeng, shuo.yang, lijinze-hku\}@connect.hku.hk,
\{zewliu, chngai\}@eee.hku.hk,
zheyu.chen@bit.edu.cn,
smyjz19@nottingham.edu.cn,
heweiw@alumni.cmu.edu,
}

\usepackage{bibentry}
\begin{document}

\maketitle

\begin{abstract}
Multiple clustering aims to discover diverse latent structures from different perspectives, yet existing methods generate exhaustive clusterings without discerning user interest, necessitating laborious manual screening. Current multi-modal solutions suffer from static semantic rigidity: predefined candidate words fail to adapt to dataset-specific concepts, and fixed fusion strategies ignore evolving feature interactions. To overcome these limitations, we propose Multi-DProxy, a novel multi-modal dynamic proxy learning framework that leverages cross-modal alignment through learnable textual proxies. Multi-DProxy introduces 1) gated cross-modal fusion that synthesizes discriminative joint representations by adaptively modeling feature interactions. 2) dual-constraint proxy optimization where user interest constraints enforce semantic consistency with domain concepts while concept constraints employ hard example mining to enhance cluster discrimination. 3) dynamic candidate management that refines textual proxies through iterative clustering feedback. Therefore, Multi-DProxy not only effectively captures a user's interest through proxies but also enables the identification of relevant clusterings with greater precision. Extensive experiments demonstrate state-of-the-art performance with significant improvements over existing methods across a broad set of multi-clustering benchmarks.
\end{abstract}

\section{Introduction}
\label{sec:introduction}
Clustering, a cornerstone of unsupervised learning, aims to uncover latent structures by grouping data based on intrinsic similarities. Traditional works rely on handcrafted features or monolithic representations \cite{macqueen1967some,ng2001spectral,caron2018deep,caron2020unsupervised}, often failing to capture the multifaceted nature of real-world data. While deep clustering works \cite{chu2024image,ouldnoughi2023clip,qian2023stable,qian2022unsupervised} have improved expressiveness, they typically produce a single partitioning, disregarding the inherent complexity of data that can be meaningfully grouped from diverse perspectives. This limitation spurred the development of multiple clustering \cite{miklautz2020deep,ren2022diversified,yao2023augdmc}, which seeks diverse partitions revealing complementary structures. However, existing works generate exhaustive clusterings without discerning user interest, necessitating laborious manual screening to identify relevant groupings—a significant practical bottleneck. Multimodal information is flooding the Internet \cite{xu2025mentor,xu2025survey}. Recent works leverage multi-modal models like CLIP \cite{radford2021learning} to align user interests (expressed as keywords, e.g., ``color") with visual representations. Recent works such as Multi-MaP \cite{yao2024multi} and Multi-Sub \cite{yao24customized} employ proxy learning, where textual prompts guide the extraction of interest-biased embeddings. Despite promising results, these solutions exhibit critical limitations:
\begin{itemize}
    \item \textbf{Static Semantic Rigidity:} Predefined candidate words (e.g., {``red", ``blue", ``green"} for ``color") fail to adapt to dataset-specific concepts, leading to misalignment when LLMs' suggestions mismatch ground-truth categories.
    \item \textbf{Inflexible Feature Fusion:} Fixed fusion strategies (e.g., concatenation or simple averaging) ignore evolving feature interactions between modalities, yielding suboptimal joint representations.
    % \item Shallow Context Modeling: CLIP’s coarse-grained alignment lacks deep reasoning for nuanced interest capture, while reliance on frozen encoders restricts adaptation to user semantics.
\end{itemize}

To overcome these deficiencies, we introduce Multi-DProxy, a novel Multi-modal Dynamic Proxy Learning framework that synergizes gated cross-modal fusion, adaptive textual proxies, and dynamic candidates to generate personalized clusterings aligned with user interest. Our core innovations address the limitations head-on:
\begin{itemize}
    \item \textbf{Gated Cross-Modal Fusion:} A hierarchical attention module with sigmoid-gated residuals dynamically recalibrates visual-textual interactions, prioritizing discriminative attributes through bidirectional feature modulation.
    \item \textbf{Dual-Constraint Proxy Optimization:} We enforce semantic consistency via user interest constraints (aligning proxies with concept centroids) while enhancing cluster discrimination via concept constraints using contrastive learning on fused features and relevant proxies. This replaces rigid candidate sets with learnable, semantically grounded proxies.
    \item \textbf{Dynamic Candidate Management:} An iterative feedback loop refines textual semantics by scoring candidates against evolving cluster centroids. This continuously adapts proxies to emergent data structures, mitigating static rigidity.
\end{itemize}

Multi-DProxy not only precisely captures user interests but also enables efficient identification of relevant clustering. Theoretical analysis proves proxy stability under dynamic updates and elucidates how visual features gate textual representations to prioritize salient attributes during fusion. Extensive experiments on a broad set of multi-clustering benchmarks demonstrate state-of-the-art performance. Our contribution can be summarized as:
\begin{itemize}
    \item The first framework unifying learnable textual proxies, dynamic candidate refinement, and adaptive feature fusion for interest-aware multiple clustering.
    \item A theoretically grounded dual-constraint mechanism ensuring semantic coherence and cluster discrimination.
    \item We conduct extensive experiments on all publicly available multiple clustering tasks, which empirically demonstrate the superiority of the proposed Multi-DProxy in precisely capturing the user’s interest.
\end{itemize}
\section{Related Work}
\label{sec:related_work}
% \subsection{Multiple Clustering}
Multiple clustering explores diverse data partitions from different perspectives, gaining increasing attention. Early methods rely on hand-crafted rules and representations. For example, COALA \cite{bae2006coala} generates new clusters using existing ones as a constraint, Hu et al. \cite{hu2017finding} maximized eigengap across subspaces, and Dang et al. \cite{dang2010generation} utilize an expectation-maximization framework to optimize mutual information. Recent approaches leverage learning-based techniques for better representations. For instance, ENRC \cite{miklautz2020deep} optimizes clustering objectives within a latent space learned by an auto-encoder, iMClusts \cite{ren2022diversified} leverages auto-encoders and multi-head attention to learn diverse feature representations, and AugDMC \cite{yao2023augdmc} applies data augmentation to generate diverse image perspectives. However, it remains challenging to identify the clustering most relevant to user interests. Recently, Multi-MaP \cite{yao2024multi} and Multi-Sub \cite{yao24customized} integrate CLIP embeddings with proxy learning to generate data representations aligned with user interests. While effective, these methods exhibit static semantic rigidity: predefined candidate words fail to adapt to dataset-specific concepts, fixed fusion strategies ignore evolving feature interactions, and CLIP inherently lacks deep contextual understanding for nuanced intent capture \cite{yao2024multi,yao24customized}. To address these limitations, we propose Multi-DProxy, a multi-modal dynamic proxy learning framework. Unlike static methods, Multi-DProxy leverages learnable textual proxies optimized via dual constraints—semantic consistency via concept centroid alignment and cluster discrimination via hard example mining. 
% It dynamically refines textual semantics through iterative clustering feedback and synthesizes joint representations using gated cross-modal fusion with hierarchical attention and adaptive recalibration. This ensures adaptability to emerging data structures and precise alignment with abstract user intents, backed by theoretical guarantees on proxy stability and cross-modal discriminability.
\section{Methodology}
\label{sec:methodology}
\begin{figure*}
    \centering
    \includegraphics[width=1\linewidth]{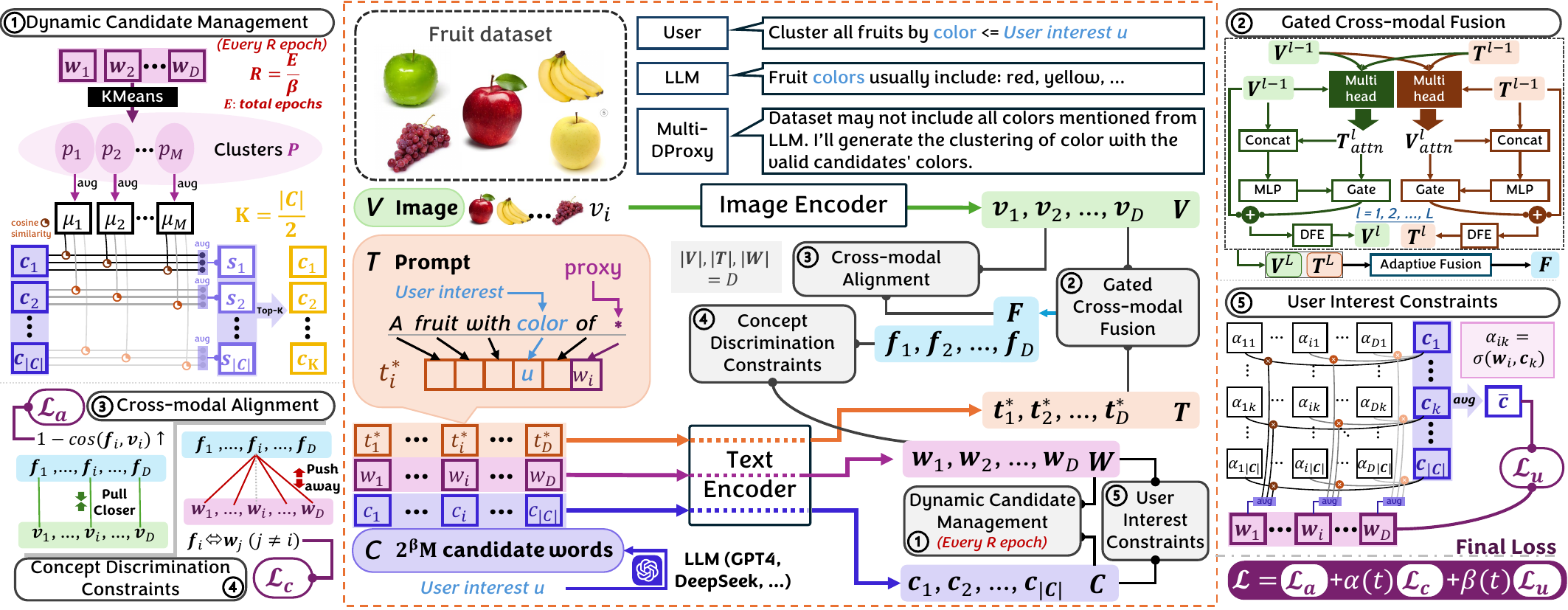}
     % \vskip -0.1in
    \caption{Overview of the Multi-DProxy framework. The central pipeline illustrates the overall architecture, while the key components are detailed on both sides: (1) Dynamic Candidate Management updates candidate words every $R$ epochs; (2) Gated Cross-modal Fusion integrates visual and textual representations; (3) Cross-modal Alignment reduces modality discrepancies; (4) Concept Discrimination Constraints enhance cluster separability; and (5) User Interest Constraints ensure alignment with domain-specific concepts.}
    \label{fig:overview}
\end{figure*}

Multi-DProxy introduces a novel dynamic proxy learning framework that generates personalized clusterings aligned with user intent through adaptive cross-modal alignment. Multi-DProxy transforms high-level concepts into learnable textual proxies that guide visual feature extraction. As illustrated in Figure~\ref{fig:overview}. 

\subsection{Multi-modal Pre-training}
First, we briefly review the training objective in CLIP as follows, and then describe the details of our Multi-DProxy method based on that. Given a set of image-text pairs as $\left\{v_i, t_i\right\}_{i=1}^D$, where $D$ is the total number of datasets, and $v_i$ is an image and $t_i$ is the corresponding text description, their vision and text representations can be obtained by two encoders as $\mathbf{v}_i=f_t(v_i) \in \mathbb{R}^d$ and $\mathbf{t}_i=f_t(t_i) \in \mathbb{R}^d$, where $\mathbf{v}_i$ and $\mathbf{t}_i$ have the unit norm and $d$ is latent dimmension. Multi-DProxy employs frozen pre-trained CLIP encoders ($f_v(\cdot)$ for vision and $f_t(\cdot)$ for text). Moreover, a user-specified concept $u$ (e.g., ``color") to refer to user interest. 

\subsection{Base Proxy Initialization}
For each input image $x_i$, we generate an initial base proxy embedding by processing a unified placeholder token ``*" using CLIP's reference word embedding function: $\mathbf{w}_i^{\prime}=f_{t}(\text{``*"}) \in \mathbb{R}^d$. We initialize and maintain $D$ different proxies $\mathbf{W}=\left\{\mathbf{w}_1, \ldots, \mathbf{w}_D\right\}$. Multi-DProxy optimizes adaptive proxy embeddings $\mathbf{w}_i$, and facilitate identifying relevant clustering through three interconnected components:
\begin{itemize} 
    \item \textbf{Gated Cross-Modal Fusion:} synthesizes discriminative joint representations through adaptive feature interaction.
     \item \textbf{Dynamic Candidate Management:} iteratively refines textual semantics via clustering feedback.
    \item \textbf{Dual-Constraint Proxy Optimization:} ensures semantic consistency while enhancing cluster discrimination.
   
    % \item 
\end{itemize}

\subsection{Gated Cross-Modal Fusion}
We propose a Gated Cross-Modal Fusion module that dynamically synthesizes discriminative joint representations through hierarchical bidirectional attention and adaptive feature recalibration. Let $\mathbf{V} = \{\mathbf{v}_1, \ldots, \mathbf{v}_D\}$ and $\mathbf{T} = \{\mathbf{t}^{*}_1, \ldots, \mathbf{t}^{*}_D\}$ denote visual and textual representations, respectively. Here $\mathbf{t}^{*}_{i}=[\mathbf{t}_i;\mathbf{w}_i]$. The component comprises core parts as following:

\subsubsection{Bidirectional Cross-Attention}
For layer $l \in {1, 2, \dots, L}$:
\begin{equation}
\begin{aligned}
& \mathbf{V}_{\mathrm{attn}}^{l}=\operatorname{MultiHead}\left(\mathbf{V}^{l-1}, \mathbf{T}^{l-1}, \mathbf{T}^{l-1}\right), \\
& \mathbf{T}_{\mathrm{attn}}^{l}=\operatorname{MultiHead}\left(\mathbf{T}^{l-1}, \mathbf{V}^{l-1}, \mathbf{V}^{l-1}\right),
\end{aligned}
\end{equation}
where $\operatorname{MultiHead}(\cdot)$ implements multi-head scaled dot-product attention.

\subsubsection{Gated Residual Fusion}
Adaptive feature recalibration via sigmoid-gated residuals:
\begin{equation}
\begin{aligned}
& \mathbf{V}^{l}=\mathbf{V}^{l-1}+\sigma\left(\mathbf{W}_g^{\mathbf{V}}\left[\mathbf{V}^{l-1} ; \mathbf{V}_{\mathrm{attn}}^{l}\right]\right) \odot \mathbf{V}_{\mathrm{attn}}^{l}, \\
& \mathbf{T}^{l}=\mathbf{T}^{l-1}+\sigma\left(\mathbf{W}_g^{\mathbf{T}}\left[\mathbf{T}^{l-1} ; \mathbf{T}_{\mathrm{attn}}^{l}\right]\right) \odot \mathbf{T}_{\mathrm{attn}}^{l},
\end{aligned}
\end{equation}
$\sigma(\cdot)$ denotes the sigmoid function (distinct from $\gamma$ in Eq.3). 
Projection matrices $\mathbf{W}_{g}^{\mathbf{V}} \in \mathbb{R}^{d \times 2d}$ 
and $\mathbf{W}_{g}^{\mathbf{T}} \in \mathbb{R}^{d \times 2d}$ transform concatenated features $[\cdot;\cdot]$.

\subsubsection{Discriminative Feature Enhancement (DFE)}
Post-attention refinement via LayerNorm and FFN:
\begin{equation}
\begin{aligned}
& \mathbf{V}^{l}=\operatorname{LayerNorm}\left(\mathbf{V}^{l}+\operatorname{FFN}\left(\mathbf{V}^{l}\right)\right), \\
& \mathbf{T}^{l}=\operatorname{LayerNorm}\left(\mathbf{T}^{l}+\operatorname{FFN}\left(\mathbf{T}^{l}\right)\right).
\end{aligned}
\end{equation}

\subsubsection{Adaptive Feature Fusion}
Final representation synthesis via temperature-scaled cosine similarity:
\begin{equation}
\mathbf{F}=\lambda \mathbf{T}^L+(1-\lambda) \mathbf{V}^L, \quad \lambda=\sigma\left(\frac{\left\langle\mathbf{T}^L, \mathbf{V}^L\right\rangle}{\tau}\right),
\end{equation}
where $\lambda \in [0,1]$ is a learnable dynamic modality weight, $\tau$ is a learnable temperature parameter (initialized by 0.1), $\langle \cdot, \cdot \rangle$ is inner product, and $\sigma(\cdot)$ is the sigmoid function. This dynamically balances modal contributions based on inter-modal agreement.

\subsection{Dynamic Candidate Management}
To overcome static semantic rigidity, we introduce a Dynamic Candidate Set that evolves with the clustering structure through iterative refinement. The system maintains and dynamically updates candidate words based on their alignment with emerging cluster structures. The update process occurs every $R$ epochs (where $R$ is a configurable update interval hyperparameter) as follows:
\begin{itemize}
    \item \textbf{Proxy Embedding Collection:} Collect all learnable proxy embeddings $\mathbf{W}=\left\{\mathbf{w}_1, \ldots, \mathbf{w}_D\right\}$ from the current training state.
    \item \textbf{Cluster Analysis:} Perform K-means clustering on the proxy embeddings to discover latent structures: $\mathbf{P} = \left\{\mathbf{p}_1, \ldots, \mathbf{p}_M\right\}=\operatorname{KMeans}(\mathbf{W}, M)$, where $M$ corresponds to the number of ground-truth classes.
    \item \textbf{Cluster Centroid Calculation:} Compute centroids for each discovered cluster: $\boldsymbol{\mu}_i=\frac{1}{\left|\mathbf{p}_i\right|} \sum_{j \in \mathbf{p}_i} \mathbf{w}_j$. 
    \item \textbf{Candidate Scoring:} Evaluate each candidate word $c_i$ by measuring its average similarity to all cluster centroids: $\mathbf{s}_i=\frac{1}{M} \sum_{j=1}^M \cos \left(\mathbf{c}_i, \boldsymbol{\mu}_j\right)$, where cosine similarity serves as the alignment metric.
    \item \textbf{Candidate Selection:} Update the candidate set by retaining the top-$K$ candidates with the highest alignment scores: $\mathcal{C}_{\text {new }}=\arg \operatorname{top-K}_{c_i \in \mathcal{C}}(\mathbf{s}_i)$, where $K = |\mathcal{C}|/2$. 
    \item \textbf{Embedding Refresh:} Recompute embeddings for the new candidate set $\mathcal{C}_{\text {new}}$ using CLIP's reference word embedding function: $\mathbf{C}_{\text {new}}=f_t\left(\mathcal{C}_{\text {new}}\right)$.
\end{itemize}
Here, the update cycle $R$ is a hyperparameter. This closed-loop refinement strategy enables continuous adaptation to emergent data patterns. The candidate set evolves from generic initial wide range concepts (e.g., {"red", "green", "blue", "burgundy", "emerald", "cyan", ...} for color) to dataset-specific semantics (e.g., {"green", "emerald", "cyan", ...}) through iterative feedback from the clustering process. 

\begin{remark}
    Initially, the LLM generates $2^\beta M$ candidate words (refer to the \textbf{Dual-Constraint Proxy Optimization} section), where $\beta=E / R, E$ represents the total number of training epochs, and $R$ denotes the interval for updating candidates. After completing $E$ epochs of training, the candidate words are reduced to $M$, aligning with the number of ground-truth classes. This ensures that clustering is not misled by erroneous guidance and effectively filters out dataset-irrelevant candidate words generated by the LLM throughout the process.
\end{remark}

\subsection{Dual-Constraint Proxy Optimization}
\subsubsection{User Interest Constraints} To enforce proxies alignment with domain concepts, we initialize candidate words $\mathcal{C}=$ $\left\{c_1, \ldots, c_{2^{\beta}M}\right\}$ using GPT-4 (e.g., \{``red", ``blue", ``green"\} for user interest $u$ ``color") with embeddings $\mathbf{C}=$ $\left\{\mathbf{c}_1, \ldots, \mathbf{c}_{2^{\beta}K}\right\} \in \mathbb{R}^{{2^{\beta}M} \times d}$, where $M$ is the total number of ground-truth classes, embedding of each candidate is computed by $\mathbf{c}_i = f_t(c_i)$. Notably, $\beta$ is a scalar parameter calculated by $E/R$, where $E$ is the total training epochs and $R$ is the configurable referring interval hyperparameter introduced in \textbf{Dynamic Candidate Management} section. These Candidate embeddings inject GPT-4's domain knowledge as semantic priors. Each proxy is computed as a semantic-weighted combination:
\begin{equation}
\mathbf{w}_i=\sum_{k=1}^{|\mathbf{C}|} \alpha_{i k} \mathbf{c}_k, \quad \alpha_{i k}=\frac{\exp \left(\mathbf{w}_i^{\prime \top} \mathbf{c}_k / \tau_{\alpha}\right)}{\sum_j^{|\mathbf{C}|} \exp \left(\mathbf{w}_i^{\prime \top} \mathbf{c}_j / \tau_{\alpha}\right)},
\end{equation}
where $\mathbf{w}_i^{\prime}$ denotes the basic proxy and $\tau_{\alpha}$ the temperature parameter. Proxies $\mathbf{w}_i$ explicitly represent weighted combinations of domain concepts $\mathbf{c}_k$. The semantic consistency loss minimizes deviation from the concept centroid:
\begin{equation}
\mathcal{L}_u=\frac{1}{D} \sum_{i=1}^D\left\|\mathbf{w}_i-\overline{\mathbf{c}}\right\|_2^2, \quad \overline{\mathbf{c}}=\frac{1}{|\mathbf{C}|} \sum_{k=1}^{|\mathbf{C}|} \mathbf{c}_k,
\end{equation}
where $\overline{\mathbf{c}} \in \mathbb{R}^d$ is the centroid of candidate embeddings. $\mathcal{L}_{u}$ ensures semantic coherence with user-specified concept $u$.

\subsubsection{Concept Discrimination Constraints} To enhance cluster separability, we employ contrastive learning on fused features $\mathbf{F}=$ $\left\{\mathbf{f}_1, \ldots, \mathbf{f}_B\right\}$ within a batch: 
\begin{equation}
\mathcal{L}_c=\frac{1}{B} \sum_{i=1}^B \log \sum_{j \neq i} \exp \left(\mathbf{f}_i^{\top} \mathbf{w}_j / \sigma\right),
\end{equation}
where $\sigma$ controls negative sample hardness and $B$ is training batch. The inner term $\sum_{j \neq i} \exp \left( \mathbf{f}_{i}^{\top} \mathbf{w}_{j} / \sigma \right)$ computes an exponential weighted sum of the similarities between sample $i$'s fused feature $\mathbf{f}_{i}$ and all proxy vectors $\mathbf{w}_{j}$ associated with clusters other than its own. Minimizing the logarithm of this sum ($\log(\cdot)$) strongly penalizes high similarity scores between $\mathbf{f}_{i}$ and incorrect proxies $\mathbf{w}_{j}$ ($j \neq i$).

\subsection{Optimization Framework}
Let $\mathbf{v}_{i}$ and $\mathbf{f}_{i}$ denote \textit{visual} and \textit{fused} features of sample $i$, respectively. The unified loss combines:
\begin{equation}
\mathcal{L}=\underbrace{\frac{1}{D} \sum_{i=1}^D\left(1-\cos \left(\mathbf{f}_i, \mathbf{v}_i\right)\right)}_{\text {Cross-modal Alignment } \mathcal{L}_a}+\alpha(t) \mathcal{L}_u+\beta(t) \mathcal{L}_c,
\end{equation}
where constraint weights following adaptive schedules: $\alpha(t)$$ = $$\min \left(0.5,0.1+0.4 \frac{t}{E}\right), \beta(t)$$ = $$0.1 \times (1-\cos (\frac{\pi t}{E}))$, where $t$ denotes current epoch and $E$ total epochs. Empirical evidence demonstrates that this dynamic scheduling design reduces the sensitivity of pre-defined hyperparameters to different datasets while achieving consistent performance advantages. This design progressively strengthens semantic constraints while maintaining stable cluster discrimination throughout training. Moreover, the cross-modal alignment term encourages the integration of multi-modal features for the same sample, thereby reducing discrepancies among different modalities. Notably, the final clusters are calculated by fused features $\mathbf{F}$.

We present pseudo-code in Algorithm~\ref{alg:multi_dproxy} to offer a clearer and more comprehensive introduction to our Multi-DProxy. Additionally, an anonymous code repository is provided in the \textbf{Supplementary Material} for further reference.

\begin{algorithm}
\caption{Multi-DProxy Framework}
\label{alg:multi_dproxy}
\begin{algorithmic}[1]
\REQUIRE 
    $\mathcal{D}$: Dataset $\{v_i,t_i\}_{i=1}^D$, $f_v(\cdot), f_t(\cdot)$: Pre-trained CLIP encoders, $u$: User interest concept (e.g., "color"), $M$: Number of ground-truth classes, $E$: Total training epochs, $R$: Candidate update interval, $K$: Initial candidate size ($K=2^{\beta}M$ where $\beta=E/R$).

\STATE \textbf{Initialize:}
\STATE Initialize $\mathbf{W} = {\mathbf{w}_1,\ldots,\mathbf{w}_D}$. 
\STATE Get $\mathbf{V} = {\mathbf{v}_1,\ldots,\mathbf{v}_D}$, where $\mathbf{v}_i = f_v(v_i)$.
\STATE Get $\mathbf{T} = {[\mathbf{t}_1;\mathbf{w}_1],\ldots,[\mathbf{t}_D;\mathbf{w}_D]}$, where $\mathbf{t}_i = f_t(t_i)$.
\STATE Generate candidates $\mathcal{C} \gets \texttt{GPT-4}(u)$ with $|\mathcal{C}|=2^{\beta}M$.
\STATE Get $\mathbf{C} \gets f_t(\mathcal{C})$.

\FOR{epoch $t=1$ to $E$}
    \STATE // \textit{Gated Cross-Modal Fusion:}
    \FOR{$l=1$ to $L$}
        \STATE $\mathbf{V}_{\textrm{attn}}^l \gets \textrm{MultiHead}(\mathbf{V}^{l-1}, \mathbf{T}^{l-1}, \mathbf{T}^{l-1})$.
        \STATE $\mathbf{T}_{\textrm{attn}}^l \gets \textrm{MultiHead}(\mathbf{T}^{l-1}, \mathbf{V}^{l-1}, \mathbf{V}^{l-1})$.
        \STATE $\mathbf{V}^l \gets \mathbf{V}^{l-1} + \sigma(\mathbf{W}_g^{\mathbf{V}}[\mathbf{V}^{l-1};\mathbf{V}_{\textrm{attn}}^l]) \odot \mathbf{V}_{\textrm{attn}}^l$.
        \STATE $\mathbf{T}^l \gets \mathbf{T}^{l-1} + \sigma(\mathbf{W}_g^{\mathbf{T}}[\mathbf{T}^{l-1};\mathbf{T}_{\textrm{attn}}^l]) \odot \mathbf{T}_{\textrm{attn}}^l$.
        \STATE $\mathbf{V}^l \gets \textrm{LayerNorm}(\mathbf{V}^l + \textrm{FFN}(\mathbf{V}^l))$.
        \STATE $\mathbf{T}^l \gets \textrm{LayerNorm}(\mathbf{T}^l + \textrm{FFN}(\mathbf{T}^l))$.
    \ENDFOR
    \STATE $\lambda \gets \sigma(\langle \mathbf{T}^L, \mathbf{V}^L \rangle / \tau)$.
    \STATE $\mathbf{F} \gets \lambda\mathbf{T}^L + (1-\lambda)\mathbf{V}^L$.
    \STATE // \textit{Dual-Constraint Proxy Optimization:}
    \FOR{$i=1$ to $D$}
        % \STATE $\mathbf{w}'_i \gets f_t(\text{``*"})$. \COMMENT{Base proxy initialization}
        \STATE $\mathbf{w}_i \gets \sum_{k=1}^{|\mathbf{C}|} \alpha_{ik} \mathbf{c}_k$.
        % \STATE Update $\mathbf{W}$.
    \ENDFOR
    \STATE $\overline{\mathbf{c}} \gets \frac{1}{|\mathbf{C}|}\sum_k \mathbf{c}_k$.
    \STATE $\mathcal{L}_u \gets \frac{1}{|\mathcal{B}|}\sum_{i} \|\mathbf{w}_i - \overline{\mathbf{c}}\|_2^2$.
    \STATE $\mathcal{L}_c \gets \frac{1}{|\mathcal{B}|}\sum_{i} \log\sum_{j\neq i}\exp(\mathbf{f}_i^\top \mathbf{w}_j / \sigma)$.
    \STATE // \textit{Optimization Framework:}
    \FOR{batch $\mathcal{B} \in \mathcal{D}$}
        \STATE $\mathcal{L}_{\textrm{align}} \gets \frac{1}{|\mathcal{B}|}\sum_{i} (1 - \cos(\mathbf{f}_i, \mathbf{v}_i))$
        \STATE $\mathcal{L} \gets \mathcal{L}_{\textrm{align}} + \alpha(t)\mathcal{L}_u + \beta(t)\mathcal{L}_c$.
        \STATE Update parameters via $\nabla\mathcal{L}$
    \ENDFOR
        
    \STATE // \textit{Dynamic Candidate Management:} 
    \IF{$t \mod R = 0$}
        \STATE $\mathbf{P} \gets \textrm{KMeans}(\mathbf{W}, M)$.
        \STATE Compute centroids $\boldsymbol{\mu}_i \gets \frac{1}{|\mathbf{p}_i|}\sum_{j\in\mathbf{p}_i}\mathbf{w}_j$.
        \FOR{each $\mathbf{c}_i \in \mathbf{C}$}
            \STATE $s_i \gets \frac{1}{M}\sum_{j=1}^M \cos(\mathbf{c}_i, \boldsymbol{\mu}_j)$
        \ENDFOR
        \STATE Get candidate words $\mathcal{C}_{\text {new}} \gets \arg \operatorname{top-K}_{c_i \in \mathcal{C}}(\mathbf{s}_i)$.
        \STATE Update candidate embeddings $\mathbf{C} \gets f_t(\mathcal{C}_{\textrm{new}})$.
    \ENDIF
\ENDFOR
\ENSURE Fused features $\mathbf{F}$ for clustering.
\end{algorithmic}
\end{algorithm}
% \subsection{}

\subsection{Theoretical Analysis}
\begin{proposition}
\label{proposition:1}
   (Proxy Stability) The dynamic candidate update reduces semantic drift by bounding proxy divergence:
   \begin{equation}
\left\|\mathbf{w}_i^{(t+1)}-\mathbf{w}_i^{(t)}\right\|_2 \leq \gamma \max \left\|\mathbf{c}_k^{(t+1)}-\mathbf{c}_k^{(t)}\right\|_2,
\end{equation}
where $\gamma=\max _i \sum_k \alpha_{i k}$ is the maximum attention mass (bounded by 1), and $\mathbf{c}_{k}^{(t)}$ denotes candidate $k$ at iteration $t$.  The bound ensures proxy stability during candidate updates.
\end{proposition}

Proposition~\ref{proposition:1} quantifies how candidate updates control semantic drift and provides theoretical justification for dynamic refinement (Proof in \textbf{Supplementary Material}). 

\begin{theorem}
\label{theorem:1}
   (Cross-modal Attention Discriminability) The gradient of the alignment loss $\mathcal{L}_{\text{align}}$ with respect to the query projection matrix $\mathbf{W}_Q$ satisfies:
\begin{equation}
\frac{\partial \mathcal{L}_{\text {align }}}{\partial \mathbf{W}_Q} \propto \sum_{i=1}^B \mathbf{v}_i \mathbf{v}_i^{\top} \mathbf{t}_i \mathbf{t}_i^{\top} \mathbf{\Lambda}_i+\mathcal{O}(\epsilon)
\end{equation}
where $\mathbf{t}_i = f_t(t_i)$ and $\mathbf{v}_i = f_v(v_i)$ denote text and visual features for the $i$-th sample. $\mathbf{\Lambda}_i=\frac{\partial \mathcal{L}_{\text {align }}}{\partial \cos \left(\mathbf{f}_i, \mathbf{v}_i\right)} \cdot \frac{1}{\left\|\mathbf{f}_i\right\|\left\|\mathbf{v}_i\right\|}$ is normalization factor, $B$ is batch size, and $\mathcal{O}(\epsilon)$ is higher-order terms. This demonstrates that visual features $\mathbf{v}_i$ modulate text representations proportionally to their discriminative power $\mathbf{v}_i\mathbf{v}_i^\top$, prioritizing semantically salient attributes during fusion.
\end{theorem}

Theorem~\ref{theorem:1} reveals visual features gate text representation learning and explains why discriminative attributes are prioritized (Proof in \textbf{Supplementary Material}).
\section{Experiment}
\label{sec:experiment}
\subsection{Dataset}
To demonstrate the effectiveness of Multi-DProxy, we conduct extensive evaluations across a diverse array of publicly available visual datasets commonly adopted for multi-clustering benchmarks \cite{yao2024multi}. This comprehensive datasets includes: \textbf{Stanford Cars} \cite{yao2024multi}, \textbf{Card}, \textbf{CMUface} \cite{gunnemann2014smvc}, \textbf{Flowers} \cite{yao2024multi}, \textbf{Fruit} \cite{hu2017finding}, \textbf{Fruit360} \cite{yao2023augdmc}, and \textbf{CIFAR-10} \cite{yao24customized}. Detailed introduction and statistical information are provided in the \textbf{Supplementary Material}.

\subsection{Baseline}
We compare our proposed Multi-DProxy approach with eight state-of-the-art multiple clustering works. These works are as follows: \textbf{MSC} \cite{hu2017finding}, \textbf{MCV} \cite{guerin2018improving}, \textbf{ENRC} \cite{miklautz2020deep}, \textbf{iMClusts} \cite{ren2022diversified}, \textbf{AugDMC} \cite{yao2023augdmc}, \textbf{DDMC} \cite{yao2024dual}, \textbf{Multi-DProxy} \cite{yao2024multi}, and \textbf{Multi-Sub} \cite{yao24customized}. Details in the \textbf{Supplementary Material}.
% \begin{itemize}
%     \item \textbf{MSC} \cite{hu2017finding}, which utilizes hand-crafted features to automatically identify distinct feature subspaces for different clustering.
%     \item \textbf{MCV} \cite{guerin2018improving}, which employs multiple pre-trained feature extractors to represent different views of the same data.
%     \item \textbf{ENRC} \cite{miklautz2020deep}, which integrates autoencoders with clustering objectives to generate diverse clustering.
%     \item \textbf{iMClusts} \cite{ren2022diversified}, which leverages the representational power of deep autoencoders and multi-head attention to produce multiple salient embedding matrices and corresponding clustering.
%     \item \textbf{AugDMC} \cite{yao2023augdmc}, which uses data augmentations to automatically extract features corresponding to various aspects of the data through a self-supervised prototype-based representation learning approach.
%     \item \textbf{DDMC} \cite{yao2024dual}, which combines disentangled representation learning with a variational Expectation-Maximization (EM) framework.
%     \item \textbf{Multi-DProxy} \cite{yao2024multi}, which relies on contrastive user-defined concepts to learn proxies tailored to user-specific interests.
%     \item \textbf{Multi-Sub} \cite{yao24customized}, which incorporates multi-modal subspace proxy learning and leverages the synergistic capabilities of CLIP and GPT-4 to better capture user preferences.
% \end{itemize}

\subsection{Metric}
To ensure the comparison's fairness, we follow the widely used settings \cite{yao2024multi,yao24customized} for evaluation. Specifically, we run k-means $10$ times and report the average clustering performance using two metrics, namely, Normalized Mutual Information (NMI) \cite{white2004performance} and Rand index (RI) \cite{rand1971objective}. These metrics range from $0$ to $1$, with higher values indicating more accurate clustering results.

\subsection{Hyperparameter}
For each user’s preference, we train the model for $E = 1000$ epochs using the Adam optimizer with a momentum of $0.9$. For all baselines, we conduct a comprehensive hyperparameter grid search based on their original papers. For our Multi-DProxy, we perform a grid search on learning rate from $\{1e^{-1}, 5e^{-2}, 1e^{-2}, 5e^{-3}, 1e^{-3}, 5e^{-4}\}$, weight decay from $\{5e^{-4}, 1e^{-4}, 5e^{-5}, 1e^{-5}, 0\}$ for all the experiments. Moreover, we tune candidate update interval $R$ from $\{100, 200, 500\}$, temperature hyperparameters $\tau_{\alpha}$ and $\sigma$ from $\{0.1, 0.2, 0.3, 0.4, 0.5\}$, respectively. We follow most previous works in obtaining each clustering by applying KMeans \cite{lloyd1982least} to the learned representations. All experiments are performed on NVIDIA RTX 4090 GPUs.

\subsection{Implementation Detail}
To ensure fairness, we adopt the same settings as the baselines \cite{yao2024multi,yao24customized} by selecting CLIP as the multi-modal model. Similarly, for LLMs, we follow the baselines \cite{yao2024multi,yao24customized} to choose GPT-4. The impact of different multi-modal models and LLMs on performance is further explored and discussed in detail in the \textbf{Ablation Study} section.

\subsection{Performance Comparison}
\begin{table*}[!t]
    \centering
    \resizebox{\linewidth}{!}{
    \begin{tabular}{c|c|cccccccccccccccc}
    \toprule
         Dataset& Metric& \multicolumn{2}{c}{Fruit} & \multicolumn{2}{c}{Fruit360} & \multicolumn{2}{c}{Card} & \multicolumn{4}{c}{CMUface} & \multicolumn{2}{c}{Standford Cars} & \multicolumn{2}{c}{Flowers} & \multicolumn{2}{c}{CIFAR-10}\\
         \midrule
         Clustering& &Color& Species& Color& Species& Order& Suits& Emotion& Glass& Identity& Pose& Color& Type& Color& Species& Type& Environment\\ \midrule
         \multirow{2}{*}{MSC}& NMI$\uparrow$& 0.6886& 0.1627& 0.2544& 0.2184& 0.0807& 0.0497& 0.1284& 0.1420& 0.3892& 0.3687& 0.2331& 0.1325& 0.2561& 0.1326& 0.1547& 0.1136\\
         & RI$\uparrow$& 0.8051& 0.6045& 0.6054& 0.5805& 0.7805& 0.3587& 0.6736& 0.5745& 0.7326& 0.6322& 0.6158& 0.5336& 0.5965& 0.5273& 0.3296& 0.3082\\ \midrule
         \multirow{2}{*}{MCV}& NMI$\uparrow$& 0.6266& 0.2733& 0.3776& 0.2985& 0.0792& 0.0430& 0.1433& 0.1201& 0.4637& 0.3254& 0.2103& 0.1650& 0.2938& 0.1561& 0.1618& 0.1379\\
         & RI$\uparrow$& 0.7685& 0.6597& 0.6791& 0.6176& 0.7128& 0.3638& 0.5268& 0.4905& 0.6247& 0.6028& 0.5802& 0.5634& 0.5860& 0.6065& 0.3305& 0.3344\\ \midrule
         \multirow{2}{*}{ENRC}& NMI$\uparrow$& 0.7103& 0.3187& 0.4264& 0.4142& 0.1225& 0.0676& 0.1592& 0.1493& 0.5607& 0.2290& 0.2465& 0.2063& 0.3329& 0.1894& 0.1826& 0.1892\\
         & RI$\uparrow$& 0.8511& 0.6536& 0.6868& 0.6984& 0.7313& 0.3801& 0.6630& 0.6209& 0.7635& 0.5029& 0.6779& 0.6217& 0.6214& 0.6195& 0.3469& 0.3599\\ \midrule
         \multirow{2}{*}{iMClusts}& NMI$\uparrow$& 0.7351& 0.3029& 0.4097& 0.3861& 0.1144& 0.0716& 0.0422& 0.1929& 0.5109& 0.4437& 0.2336& 0.1963& 0.3169& 0.1887& 0.2040& 0.1920\\
         & RI$\uparrow$& 0.8632& 0.6743& 0.6841& 0.6732& 0.7658& 0.3715& 0.5932& 0.5627& 0.8260& 0.6114& 0.6552& 0.5643& 0.6127& 0.6077& 0.3695& 0.3664\\ \midrule
         \multirow{2}{*}{AugDMC}& NMI$\uparrow$& 0.8517& 0.3546& 0.4594& 0.5139& 0.1440& 0.0873& 0.0161& 0.1039& 0.5875& 0.1320& 0.2736& 0.2364& 0.3556& 0.1996& 0.2855& 0.2927\\
         & RI$\uparrow$& 0.9108& 0.7399& 0.7392& 0.7430& 0.8267& 0.4228& 0.5367& 0.5361& 0.8334& 0.5517& 0.7525& 0.7356& 0.6931& 0.6227& 0.4516& 0.4689\\ \midrule
         \multirow{2}{*}{DDMC}& NMI$\uparrow$& 0.8973& 0.3764& 0.4981& 0.5292& 0.1563& 0.0933& 0.1726& 0.2261& 0.6360& 0.4526& 0.6899& 0.6045& 0.6327& 0.6148& 0.3991& 0.3782\\
         & RI$\uparrow$& 0.9383& 0.7621& 0.7472& 0.7703& 0.8326& 0.6469& 0.7593& 0.7663& 0.8907& 0.7904& 0.8765& 0.7957& 0.7887& 0.8321& 0.5827& 0.5547\\ \midrule
         \multirow{2}{*}{Multi-MaP}& NMI$\uparrow$& 0.8619& \textbf{1.0000}& 0.6239& 0.5284& 0.3653& 0.2734& 0.1786& 0.3402& 0.6625& 0.4693& 0.7360& 0.6355& 0.6426& 0.6013& 0.4969& 0.4598\\
         & RI$\uparrow$& 0.9526& \textbf{1.0000}& 0.8243& 0.7582& 0.8587& 0.7039& 0.7105& 0.7068& 0.9496& 0.6624& 0.9193& 0.8399& 0.7984& 0.8103& 0.7104& 0.6737\\ \midrule
         \multirow{2}{*}{Multi-Sub}& NMI$\uparrow$& \underline{0.9693}& \textbf{1.0000}& \underline{0.6654}& \underline{0.6123}& \underline{0.3921}& \underline{0.3104}& \underline{0.2053}& \underline{0.4870}& \underline{0.7441}& \underline{0.5923}& \underline{0.7533}& \underline{0.6616}& \underline{0.6940}& \underline{0.6724}& \underline{0.5271}& \underline{0.4828}\\
         & RI$\uparrow$& \underline{0.9964}& \textbf{1.0000}& \underline{0.8821}& \underline{0.8504}& \underline{0.8842}& \underline{0.7941}& \underline{0.8527}& \underline{0.8324}& \underline{0.9834}& \underline{0.8736}& \underline{0.9387}& \underline{0.8792}& \underline{0.8843}& \underline{0.8719}& \underline{0.7394}& \underline{0.7096}\\ \midrule
         \multirow{2}{*}{Multi-DProxy}& NMI$\uparrow$& \textbf{1.0000}& \textbf{1.0000}& \textbf{0.7058}& \textbf{0.6490}& \textbf{0.5319}& \textbf{0.5008}& \textbf{0.2189}& \textbf{0.7739}& \textbf{0.7609}& \textbf{0.6646}& \textbf{0.7610}& \textbf{0.6829}& \textbf{0.7101}& \textbf{0.6888}& \textbf{0.5863}& \textbf{0.5431}\\
         & RI$\uparrow$& \textbf{1.0000}& \textbf{1.0000}& \textbf{0.8855}& \textbf{0.8537}& \textbf{0.9101}& \textbf{0.8848}& \textbf{0.8548}& \textbf{0.8381}& \textbf{0.9849}& \textbf{0.8991}& \textbf{0.9403}& \textbf{0.8901}& \textbf{0.8939}& \textbf{0.8897}& \textbf{0.7684}& \textbf{0.7204}\\ 
         \bottomrule
    \end{tabular}
    }
     % \vskip -0.1in
    \caption{Comparison with state-of-the-art methods across multiple clustering benchmarks.}
    \label{tab:performance}
\end{table*}
Table~\ref{tab:performance} presents the clustering results, showing that Multi-DProxy consistently outperforms all baselines, highlighting its superiority. This demonstrates the strong generalization ability of the pre-trained multi-modal model, which effectively captures data features from diverse perspectives. Since our method employs the multi-modal encoder and LLM to generate clustering results, a natural question arises: \textbf{how would the performance compare if they were used directly in a zero-shot manner? }

\begin{table}[!t]
    \centering
    % \small
    \resizebox{\linewidth}{!}{
    \begin{tabular}{c|c|cccccc}
    \toprule
    Dataset& Clustering& \multicolumn{2}{c}{CLIP$_\text{GPT}$}& \multicolumn{2}{c}{CLIP$_\text{label}$}& \multicolumn{2}{c}{Multi-DProxy}\\\midrule
    Metric& & NMI$\uparrow$& RI$\uparrow$& NMI$\uparrow$& RI$\uparrow$& NMI$\uparrow$& RI$\uparrow$ \\ \midrule
    \multirow{2}{*}{Fruit}& Color& 0.7912& 0.9075& 0.8629& 0.9780& \textbf{1.0000}& \textbf{1.0000} \\
    & Species& 0.9793& 0.9919& \textbf{1.0000}& \textbf{1.0000}& \textbf{1.0000}& \textbf{1.0000} \\ \midrule
    \multirow{2}{*}{Fruit360}& Color& 0.5613& 0.7305& 0.5746& 0.7673& \textbf{0.7058}& \textbf{0.8855} \\
    & Species& 0.4370& 0.7552& 0.5364& 0.7631& \textbf{0.6490}& \textbf{0.8537} \\ \midrule
    \multirow{2}{*}{Card}& Order& 0.3518& 0.8458& 0.3518& 0.8458& \textbf{0.5319}& \textbf{0.9101} \\
    & Suits& 0.2711& 0.6123& 0.2711& 0.6123& \textbf{0.5008}& \textbf{0.8848} \\
    \midrule
    \multirow{2}{*}{Card}& Order& 0.3518& 0.8458& 0.3518& 0.8458& \textbf{0.5319}& \textbf{0.9101} \\
    & Suits& 0.2711& 0.6123& 0.2711& 0.6123& \textbf{0.5008}& \textbf{0.8848} \\  \midrule
    \multirow{4}{*}{CMUface}& Emotion& 0.1576& 0.6532& 0.1590& 0.6619& \textbf{0.2189}& \textbf{0.8548} \\
    & Glass& 0.2905& 0.6869& 0.4686& 0.7505& \textbf{0.7739}& \textbf{0.8381} \\
    & Identity& 0.1998& 0.6388& 0.2677& 0.7545& \textbf{0.7609}& \textbf{0.9849}\\
    & Pose& 0.4088& 0.6473& 0.4691& 0.6409& \textbf{0.6646}& \textbf{0.8991}\\ \midrule
    \multirow{2}{*}{Stanford Cars}& Color& 0.6539& 0.8237& 0.6830& 0.8642& \textbf{0.7610}& \textbf{0.9403} \\
    & Type& 0.6207& 0.7931& 0.6429& 0.8456& \textbf{0.6829}& \textbf{0.8901} \\ \midrule
    \multirow{2}{*}{Flowers}& Color& 0.5653& 0.7629& 0.5828& 0.7836& \textbf{0.7101}& \textbf{0.8939} \\
    & Species& 0.5620& 0.7553& 0.6019& 0.7996& \textbf{0.6888}& \textbf{0.8897} \\ \midrule
    \multirow{2}{*}{CIFAR-10}& Type& 0.4935& 0.6741& 0.5087& 0.7102& \textbf{0.5863}& \textbf{0.7684} \\
    & Environment& 0.4302& 0.6507& 0.4643& 0.6801& \textbf{0.5431}& \textbf{0.7204} \\ 
    \bottomrule
    \end{tabular}
    }
     % \vskip -0.1in
    \caption{Zero-shot performance comparison.}
    \label{tab:zero_shot}
\end{table}

% More discussions of the zero-shot performance for various combinations of multi-modal models and LLMs are provided in the \textbf{Supplementary Material}.
\begin{figure}[!h]
    \centering
     % \vskip -0.05in
    \includegraphics[width=1\linewidth]{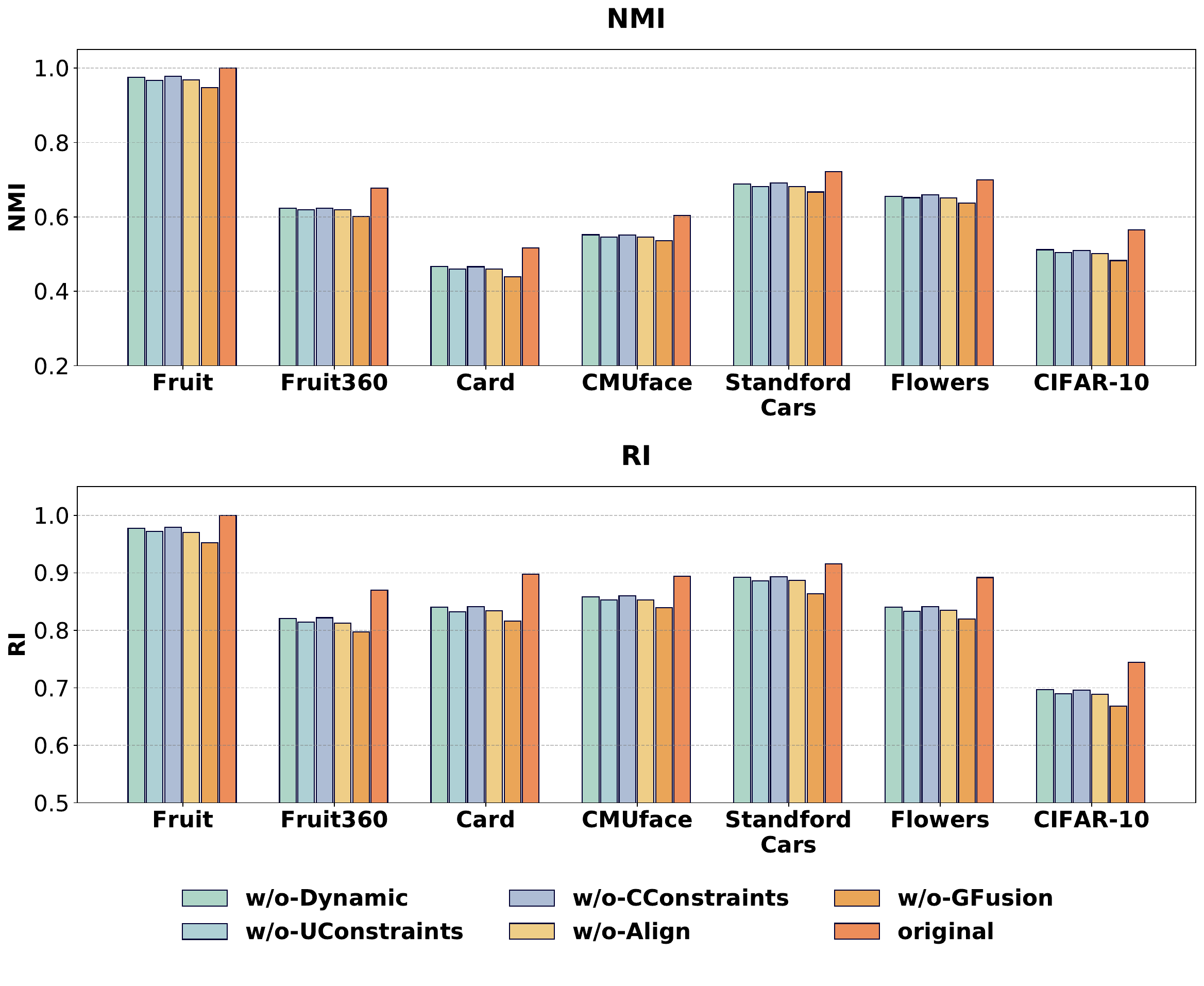}
     % \vskip -0.1in
    \caption{Ablation study. For each dataset, the average performance across all clustering objects is reported.}
    \label{fig:ablation_study}
\end{figure}

\begin{figure*}[!h]
    \centering
     % \vskip -0.05in
    \includegraphics[width=1\linewidth]{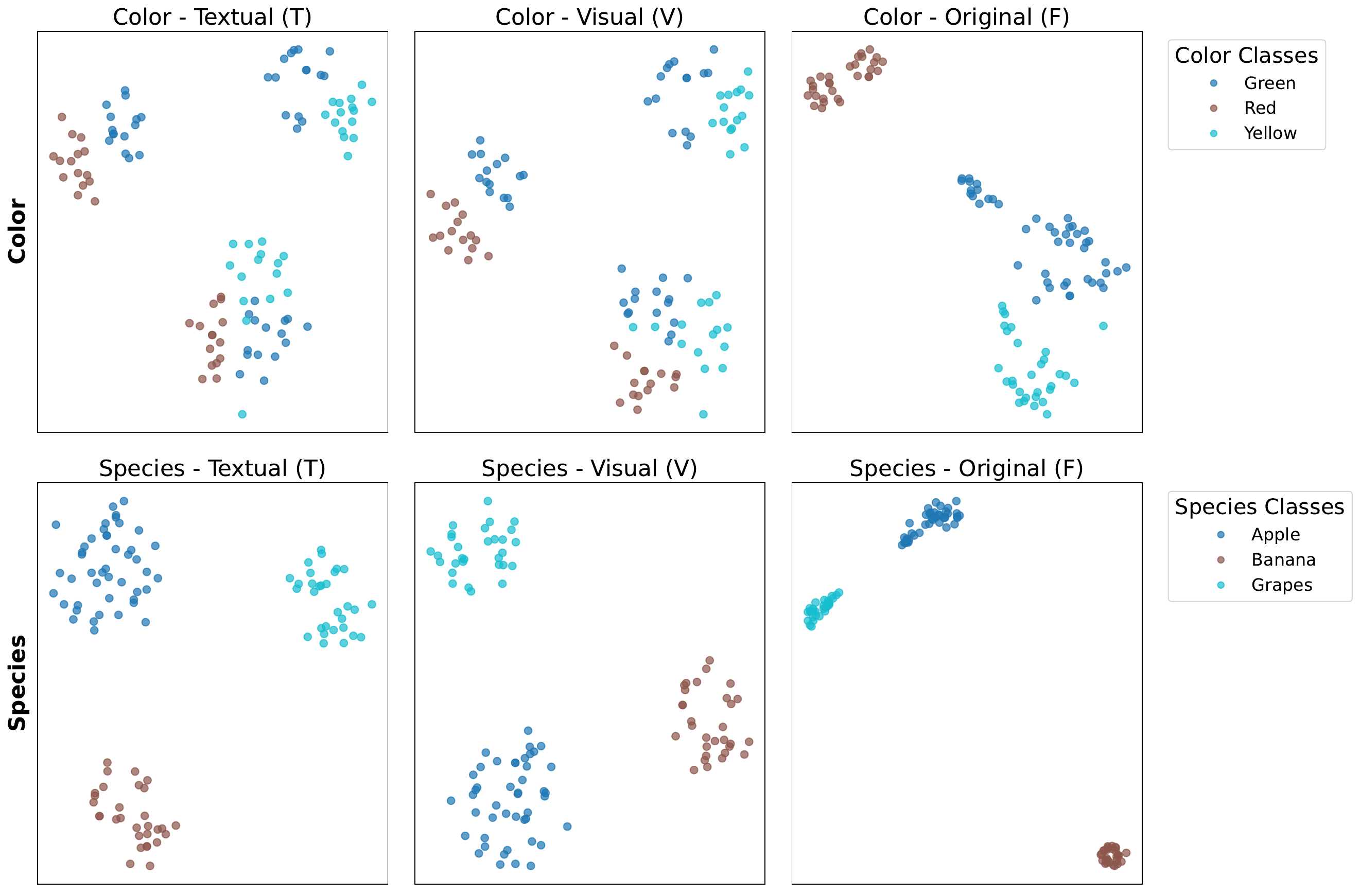}
     % \vskip -0.1in
    \caption{Visualization of textual, visual, and fused representations on the Fruit dataset.}
    \label{fig:vis}
\end{figure*}

To explore this, we introduce two zero-shot variants of CLIP: (1) \textbf{CLIP$_\text{GPT}$}, which uses GPT-4 to generate candidate labels and performs zero-shot classification with these labels as class names, and (2) \textbf{CLIP$_\text{label}$}, which directly uses the ground truth label set for zero-shot classification. Note that CLIP$_\text{label}$ leverages an unfair setting, as the ground truth labels are known in advance, providing an upper bound for CLIP's zero-shot performance. The results, shown in Table~\ref{tab:zero_shot}, align with expectations: CLIP$_\text{label}$ generally outperforms CLIP$_\text{GPT}$ due to the fixed and accurate ground truth labels, whereas CLIP$_\text{GPT}$ relies on candidate labels that may introduce noise. Notably, both methods achieve identical performance on the Cards dataset, as GPT-4 generates candidate labels perfectly matching the ground truth.

Furthermore, Multi-DProxy outperforms CLIP$_\text{GPT}$ in nearly all cases, indicating that the proposed method learns more effective features through its training process. Even when compared to CLIP$_\text{label}$, which benefits from access to the ground truth, Multi-DProxy achieves superior results in certain cases, such as clustering by color in the Fruit360 dataset. This is because CLIP tends to emphasize features from a single aspect, whereas Multi-DProxy learns a more comprehensive embedding of diverse features by leveraging user-supervised training. Additionally, Multi-DProxy achieves competitive performance relative to CLIP$_\text{label}$ in other cases, further validating the effectiveness of our proposed Multi-DProxy.

\subsection{Ablation Study}
To validate the effectiveness of our Multi-DProxy, we conduct experiments to justify the importance of key components. We design the following variants: 1) \emph{w/o-Dynamic}, which removes Dynamic Candidate Management component and directly generates $M$ candidate words via LLM. 2) \emph{w/o-UConstraints}, which removes User Interest Constraints component. 3) \emph{w/o-CConstraints}, which removes Concept Discrimination Constraints component. 4) \emph{w/o-GFusion}, which replaces Gated Cross-Modal Fusion component by directly concatenating visual and textual representations. For each dataset, the average performance across all clustering objects is reported. The results in Figure~\ref{fig:ablation_study} show that the removal of any component results in a performance drop, demonstrating the effectiveness of all components. Furthermore, the Gated Cross-Modal Fusion component has a more significant impact on the model's performance. Therefore, we conducted further exploration of the individual contributions of each modality.

We design the following variants: 1) \emph{-T}, which only uses textual modality. 2) \emph{-V}, which only uses visual modality \footnote{The ablation study evaluates the visual, textual, and fused representations of Multi-DProxy after 1000 training epochs. Thanks to the Cross-Modality Alignment component, the visual representation acquires clustering capabilities aligned with user interests.}. For each dataset, the average performance across all clustering objects is reported. The results in Figure~\ref{tab:ablation_study} demonstrate that each modality possesses the capability to perform clustering independently. Moreover, the fused representation, enhanced by our tailored modality aggregation and alignment tasks, achieves significantly superior performance. 
\begin{table}[!t]
\small
    \centering
    % \resizebox{\linewidth}{!}{
     % \vskip -0.05in
    \begin{tabular}{c|c|ccc}
    \toprule
    Variant& Metric& \emph{-T} & \emph{-V} & Original \\ \midrule
    \multirow{2}{*}{Fruit}& NMI$\uparrow$& 0.7639& 0.7421& \textbf{1.0000}\\
    & RI$\uparrow$& 0.7719& 0.7471& \textbf{1.0000}\\ \midrule
    \multirow{2}{*}{Fruit360}& NMI$\uparrow$& 0.5439& 0.5326& \textbf{0.6774}\\
    & RI$\uparrow$& 0.7410& 0.7321& \textbf{0.8696}\\ \midrule
    \multirow{2}{*}{Card}& NMI$\uparrow$& 0.4439& 0.4312& \textbf{0.5164}\\
    & RI$\uparrow$& 0.8219& 0.8138& \textbf{0.8975}\\ \midrule
    \multirow{2}{*}{CMUface}& NMI$\uparrow$& 0.5322& 0.5233& \textbf{0.6046}\\
    & RI$\uparrow$& 0.8231& 0.8150& \textbf{0.8942}\\ \midrule
    \multirow{2}{*}{Stanford Cars}& NMI$\uparrow$& 0.6459& 0.6378& \textbf{0.7220}\\
    & RI$\uparrow$& 0.8199& 0.8120& \textbf{0.9152}\\ \midrule
    \multirow{2}{*}{Flowers}& NMI$\uparrow$& 0.6369& 0.6248& \textbf{0.6995}\\
    & RI$\uparrow$& 0.8245& 0.8129& \textbf{0.8918}\\ \midrule
    \multirow{2}{*}{CIFAR-10}& NMI$\uparrow$& 0.5030& 0.4925& \textbf{0.5647}\\
    & RI$\uparrow$& 0.6875& 0.6789& \textbf{0.7444}\\ 
    \bottomrule
    \end{tabular}
    % }
     % \vskip -0.1in
    \caption{Ablation study. For each dataset, the average performance across all clustering objects is reported.}
    \label{tab:ablation_study}
\end{table}

To further demonstrate the effectiveness of the fused representation, we visualize the representations obtained for \emph{-T}, \emph{-V}, and Original Multi-DProxy. The results are shown in Figure~\ref{fig:vis}. Using visual information alone fails to achieve satisfactory clustering performance, whereas the fused representation, combining visual and textual information, better aligns the clustering results with user interests.

\begin{figure*}[!h]
    \centering
     % \vskip -0.05in
    \includegraphics[width=1\linewidth]{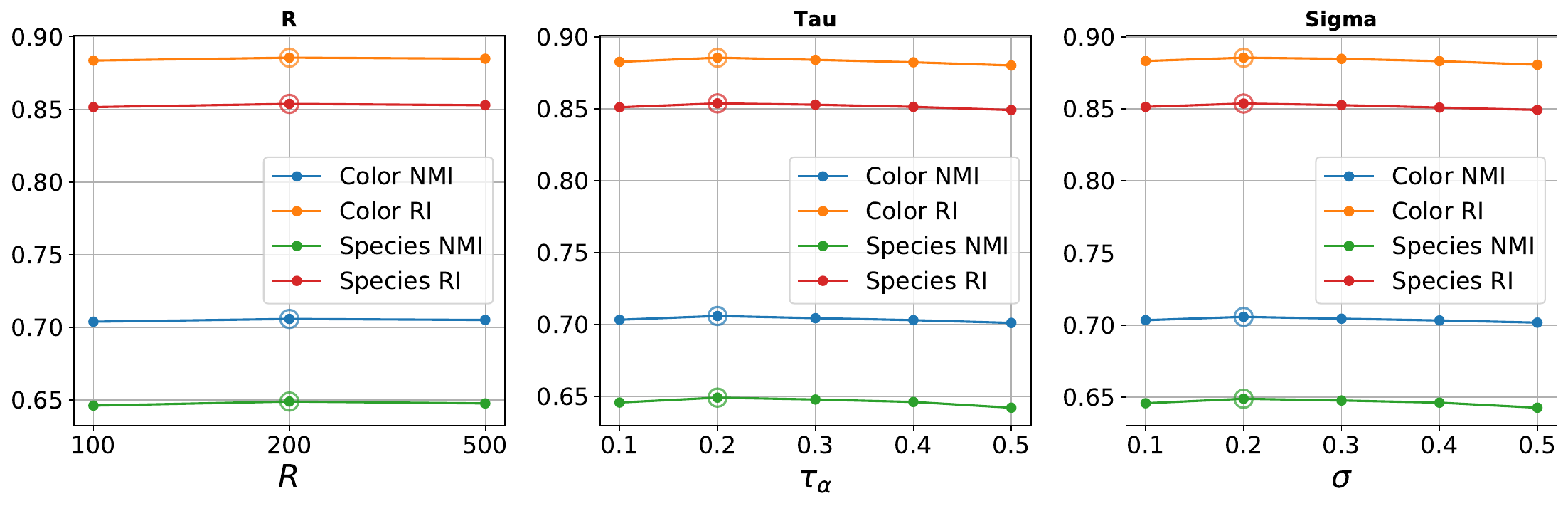}
     % \vskip -0.1in
    \caption{Hyperparameter analysis on the Fruit dataset.}
    \label{fig:hyper}
\end{figure*}

\subsection{In-depth Analysis}
We investigate the impact of different multi-modal models and LLMs on Multi-DProxy's performance. For multi-modal models, we select CLIP, ALIGN \cite{jia2021scaling}, and BLIP2 \cite{li2023blip}, with ALIGN and BLIP2 offering larger datasets and stronger representation capabilities than CLIP. For LLMs, we choose GPT4, GPT4o \cite{yang2023dawn}, and DeepSeekV3 \cite{liu2024deepseek}, with DeepSeekV3 and GPT4o providing richer knowledge than GPT4. We selected two representative datasets, and the experimental results in Table~\ref{tab:depth} show that different LLMs have a minimal impact on performance, as they are only used for generating candidate words. In contrast, stronger multi-modal models further enhance the performance of Multi-DProxy.

\begin{figure}[!h]
    \centering
     % \vskip -0.05in
    \includegraphics[width=1\linewidth]{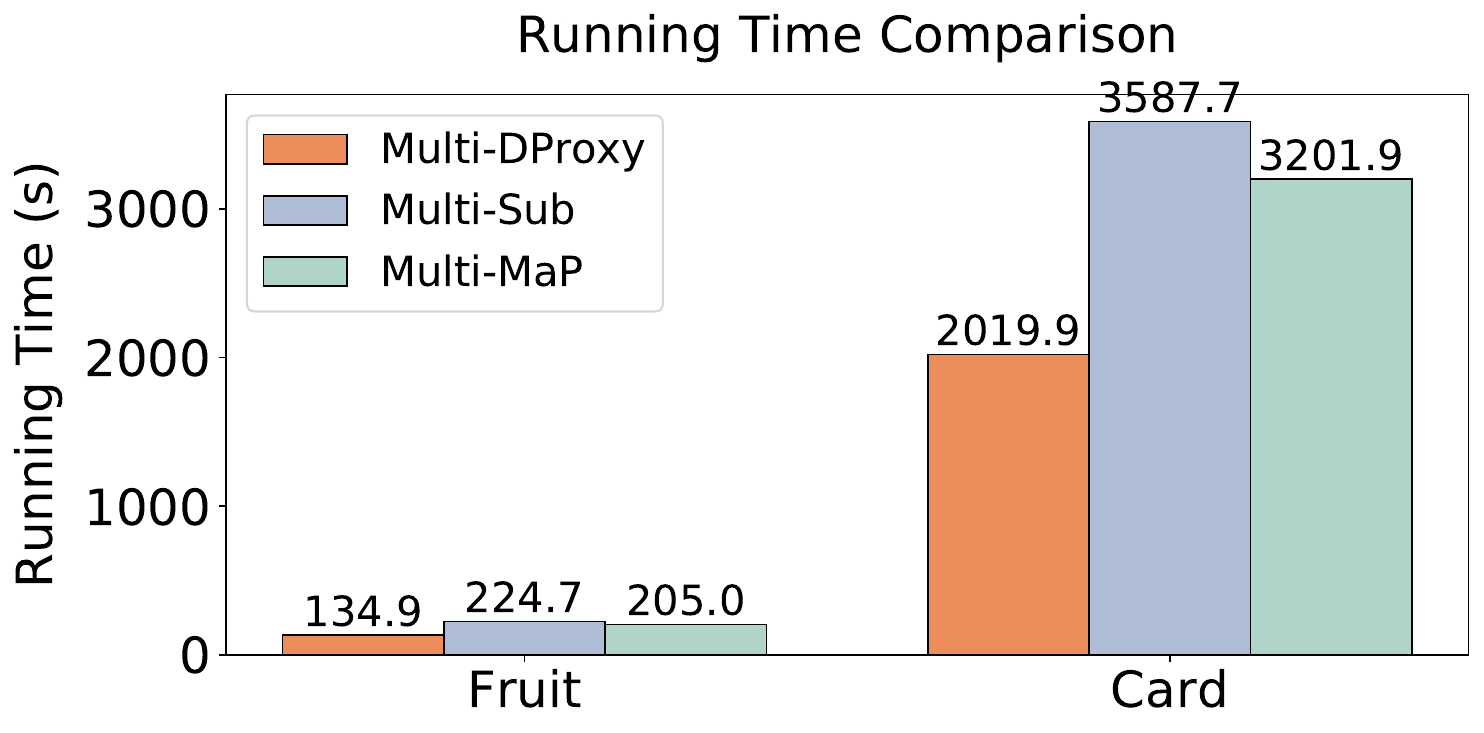}
     % \vskip -0.1in
    \caption{Efficiency study on the Fruit and Card datasets.}
    \label{fig:efficiency}
\end{figure}

We further explored the efficiency advantages of Multi-DProxy over the sub-optimal baselines (Multi-Sub and Multi-MaP). We reported the average running time for clustering objects across two datasets. As shown in Figure~\ref{fig:efficiency}, our method achieves significantly higher efficiency compared to these sub-optimal baselines.

\begin{table}[!t]
    \centering
     % \vskip -0.05in
    \resizebox{\linewidth}{!}{
    \begin{tabular}{c|c|cccccccccc}
    \toprule
         Dataset& Metric& \multicolumn{2}{c}{Fruit360} & \multicolumn{2}{c}{Card}\\
         \midrule
         Clustering& & Color& Species& Order& Suits\\ \midrule
         \multirow{2}{*}{CLIP-GPT4}& NMI$\uparrow$& 0.7058& 0.6490& 0.5319& 0.5008\\
         & RI$\uparrow$& 0.8855& 0.8537& 0.9101& 0.8848\\ 
         \midrule
         \multirow{2}{*}{CLIP-GPT4o}& NMI$\uparrow$& 0.7059& 0.6487& 0.5323& 0.5014\\
         & RI$\uparrow$& 0.8862& 0.8539& 0.9109& 0.8860\\ 
         \midrule
         \multirow{2}{*}{CLIP-DeepSeekV3}& NMI$\uparrow$& 0.7048& 0.6451& 0.5322& 0.5006\\
         & RI$\uparrow$& 0.8850& 0.8516& 0.9088& 0.8829\\ 
         \midrule
         \multirow{2}{*}{ALIGN-GPT4}& NMI$\uparrow$& 0.7289& 0.6647& 0.5809& 0.5215\\
         & RI$\uparrow$& 0.8998& 0.8717& 0.9218& 0.8901\\ 
         \midrule
         \multirow{2}{*}{ALIGN-GPT4o}& NMI$\uparrow$& \textbf{0.7300}& 0.6655& 0.5815& \textbf{0.5224}\\
         & RI$\uparrow$& \textbf{0.9007}& \textbf{0.8728}& \textbf{0.9223}& \textbf{0.8911}\\ 
         \midrule
         \multirow{2}{*}{ALIGN-DeepSeekV3}& NMI$\uparrow$& 0.7292& \textbf{0.6661}& \textbf{0.5820}& 0.5219\\
         & RI$\uparrow$& 0.8995& 0.8725& 0.9219& 0.8904\\ 
         \midrule
         \multirow{2}{*}{BLIP2-GPT4}& NMI$\uparrow$& 0.7281& 0.6597& 0.5628& 0.5178\\
         & RI$\uparrow$& 0.8995& 0.8699& 0.9190& 0.8872\\ 
         \midrule
         \multirow{2}{*}{BLIP2-GPT4o}& NMI$\uparrow$& 0.7268& 0.6600& 0.5641& 0.5183\\
         & RI$\uparrow$& 0.9002& 0.8711& 0.9197& 0.8881\\ 
         \midrule
         \multirow{2}{*}{BLIP2-DeepSeekV3}& NMI$\uparrow$& 0.7285& 0.6592& 0.5629& 0.5191\\
         & RI$\uparrow$& 0.9000& 0.8715& 0.9188& 0.8874\\ 
         \bottomrule
    \end{tabular}
    }
     % \vskip -0.1in
    \caption{Performance comparison across different multimodal models and LLMs.}
    \label{tab:depth}
\end{table}

\subsection{Hyperparameter Analysis}
We further investigate the effect of candidate update interval $R$, temperature hyperparameter $\tau_\alpha$, and hyperparameter $\sigma$, respectively. We provide results on the Fruit360 dataset. Figure~\ref{fig:hyper} shows that the optimal choices for the temperature hyperparameters $\tau_\alpha$ and $\sigma$ are both 0.2, while the optimal choice for the candidate update interval $R$ is 200. Notably, the model's hyperparameters exhibit good performance across a reasonable range.

\section{Conclusion}
In this work, we presented Multi-DProxy, a dynamic proxy learning framework that overcomes the semantic rigidity of existing multi-modal clustering methods. By integrating learnable textual proxies refined through clustering feedback and gated cross-modal fusion that prioritizes discriminative features, our approach achieves precise alignment with user interests. Extensive validation across a board set of benchmarks shows state-of-the-art performance.

\section*{Acknowledgements}
This work was supported by the Hong Kong UGC General Research Fund no. 17203320 and 17209822, and the project grants from the HKU-SCF FinTech Academy.

% \bigskip
% \noindent Thank you for reading these instructions carefully. We look forward to receiving your electronic files!

% \bibliographystyle{IEEEtran}
% \bibliographystyle{aaai}
% \bibliography{main}

\bibliography{aaai2026}

\clearpage
\appendix
\section{Appendix}
\subsection{Proof of Proposition~\ref{proposition:1}}
Consider the proxy formulation at iteration $t$:
\begin{equation}
\mathbf{w}_i^{(t)}=\sum_{k=1}^K \alpha_{i k}^{(t)} \mathbf{c}_k^{(t)}.
\end{equation}
After candidate update at $t+1$:
\begin{equation}
\mathbf{w}_i^{(t+1)}=\sum_{k=1}^K \alpha_{i k}^{(t+1)} \mathbf{c}_k^{(t+1)}.
\end{equation}
The difference is bounded by:
\begin{equation}
\begin{aligned}
&\left\|\mathbf{w}_i^{(t+1)}-\mathbf{w}_i^{(t)}\right\|_2  =\left\|\sum_k \alpha_{i k}^{(t+1)} \mathbf{c}_k^{(t+1)}-\sum_k \alpha_{i k}^{(t)} \mathbf{c}_k^{(t)}\right\|_2 \\
& \leq \underbrace{\left\|\sum_k \alpha_{i k}^{(t+1)}\left(\mathbf{c}_k^{(t+1)}-\mathbf{c}_k^{(t)}\right)\right\|_2}_{\text {Term A }}+\underbrace{\left\|\sum_k\left(\alpha_{i k}^{(t+1)}-\alpha_{i k}^{(t)}\right) \mathbf{c}_k^{(t)}\right\|_2}_{\text {Term B }}.
\end{aligned}
\end{equation}
\noindent \textbf{Bounding Term A:} Since $\alpha_{i k}$ are convex weights ( $\sum_k \alpha_{i k}=1, \alpha_{i k} \geq 0$ ):
\begin{equation}
\begin{aligned}
\operatorname{Term} \mathrm{A} &\leq \sum_k \alpha_{i k}^{(t+1)}\left\|\mathbf{c}_k^{(t+1)}-\mathbf{c}_k^{(t)}\right\|_2 \\&\leq\left(\sum_k \alpha_{i k}^{(t+1)}\right) \max _k\left\|\mathbf{c}_k^{(t+1)}-\mathbf{c}_k^{(t)}\right\|_2.
\end{aligned}
\end{equation}
\noindent \textbf{Bounding Term B:} From the softmax formulation:
\begin{equation}
\left|\alpha_{i k}^{(t+1)}-\alpha_{i k}^{(t)}\right| \leq L_\alpha\left\|\mathbf{w}_i^{(t+1)}-\mathbf{w}_i^{(t)}\right\|_2,
\end{equation}
where $L_\alpha$ is the Lipschitz constant of softmax. Since candidate embeddings are bounded ( $\left\|\mathbf{c}_k\right\| \leq$ $M)$:
\begin{equation}
\operatorname{Term} \mathrm{B} \leq M \sum\left|\alpha_{i k}^{(t+1)}-\alpha_{i k}^{(t)}\right| \leq M K L_\alpha\left\|\mathbf{w}_i^{(t+1)}-\mathbf{w}_i^{(t)}\right\|_2,
\end{equation}
where $K$ denotes the number of candidate words and $M$ denotes the number of cores for K-means. Combining both bounds:
\begin{equation}
\left\|\mathbf{w}_i^{(t+1)}-\mathbf{w}_i^{(t)}\right\|_2 \leq \gamma \max _k\left\|\mathbf{c}_k^{(t+1)}-\mathbf{c}_k^{(t)}\right\|_2+\kappa\left\|\mathbf{w}_i^{(t+1)}-\mathbf{w}_i^{(t)}\right\|_2,
\end{equation}
where $\gamma=\sum_k \alpha_{i k}^{(t+1)}$ and $\kappa=M K L_\alpha$. For sufficiently small $\kappa<1$ (ensured by normalization):
\begin{equation}
\left\|\mathbf{w}_i^{(t+1)}-\mathbf{w}_i^{(t)}\right\|_2 \leq \frac{\gamma}{1-\kappa} \max _k\left\|\mathbf{c}_k^{(t+1)}-\mathbf{c}_k^{(t)}\right\|_2.
\end{equation}
Taking $\gamma=\max _i \sum_k \alpha_{i k}=1$ completes the proof.

\subsection{Proof of Theorem~\ref{theorem:1}}
The alignment loss is defined as:
\begin{equation}
\mathcal{L}_{\text {align }}=\frac{1}{B} \sum_{i=1}^B\left(1-\cos \left(\mathbf{F}_i, \mathbf{v}_i\right)\right),
\end{equation}
where $\cos (\mathbf{a}, \mathbf{b})$$=$$\frac{\mathbf{a}^{\top} \mathbf{b}}{\|\mathbf{a}\|\|\mathbf{b}\|}$ and $\mathbf{F}_i$ is the fused representation from gated cross-modal fusion (Eq. 4).
The attention output (Eq. 1) is:
\begin{equation}
\mathbf{T}_{\text {attn }}=\operatorname{MultiHead}(\mathbf{T}, \mathbf{V}, \mathbf{V}).
\end{equation}
For single-head attention (generalizable to multi-head):
\begin{equation}
\begin{gathered}
\mathbf{Q}=\mathbf{T W}_Q, \quad \mathbf{K}=\mathbf{V} \mathbf{W}_K, \quad \mathbf{V}_{\text {val }}=\mathbf{V} \mathbf{W}_V, \\
\mathbf{T}_{\text {attn }}=\operatorname{Softmax}\left(\frac{\mathbf{Q} \mathbf{K}^{\top}}{\sqrt{d_k}}\right) \mathbf{V}_{\text {val}},
\end{gathered}
\end{equation}
The gradient decomposes as:
\begin{equation}
\frac{\partial \mathcal{L}_{\text {align }}}{\partial \mathbf{W}_Q}=\sum_{i=1}^B \frac{\partial \mathcal{L}_{\text {align }}}{\partial \cos _i} \cdot \frac{\partial \cos _i}{\partial\left(\mathbf{F}_i^{\top} \mathbf{v}_i\right)} \cdot \frac{\partial\left(\mathbf{F}_i^{\top} \mathbf{v}_i\right)}{\partial \mathbf{W}_Q},
\end{equation}
where $\frac{\partial \mathcal{L}_{\text {align }}}{\partial \cos _i}$$ = $$-\frac{1}{B}$, $\frac{\partial \cos _i}{\partial\left(\mathbf{F}_i^{\top} \mathbf{v}_i\right)}$$ = $$\mathbf{\Lambda}_i$, and $\frac{\partial(\mathbf{F}_{i}^{\top}\mathbf{v}_{i})}{\partial\mathbf{W}_{Q}}$$ = $$\mathbf{v}_{i}^{\top} \frac{\partial\mathbf{F}_{i}}{\partial\mathbf{W}_{Q}}$.
Using residual connection (Eq. 2):
\begin{equation}
\begin{gathered}
\mathbf{F}_i \approx \mathbf{t}_i+\mathbf{T}_{\mathrm{attn}, i}, \\
\frac{\partial \mathbf{F}_i}{\partial \mathbf{W}_Q} \approx \frac{\partial}{\partial \mathbf{W}_Q}\left(\sum_{j=1}^B a_{i j} \mathbf{v}_j \mathbf{W}_V\right),
\end{gathered}
\end{equation}
where $a_{i j}=\operatorname{Softmax}\left(\frac{\mathbf{t}_i \mathbf{W}_Q\left(\mathbf{v}_j \mathbf{W}_K\right)^{\top}}{\sqrt{d_k}}\right)$.
Then, we got:
\begin{equation}
\frac{\partial a_{i j}}{\partial \mathbf{W}_Q}=\sum_{k=1}^B \frac{\partial a_{i j}}{\partial s_{i k}} \frac{\partial s_{i k}}{\partial \mathbf{W}_Q},
\end{equation}
where $s_{i k}=\frac{1}{\sqrt{d_k}}\left(\mathbf{t}_i \mathbf{W}_Q\right)\left(\mathbf{v}_k \mathbf{W}_K\right)^{\top}$, $\frac{\partial a_{i j}}{\partial s_{i k}}=a_{i j}\left(\delta_{j k}-a_{i k}\right)$, and $\frac{\partial s_{i k}}{\partial \mathbf{W}_Q}=\frac{1}{\sqrt{d_k}} \mathbf{t}_i^{\top} \otimes\left(\mathbf{v}_k \mathbf{W}_K\right)$.
Under diagonal-dominant attention $\left(a_{i i} \gg a_{i j}, j \neq i\right)$:
\begin{equation}
\frac{\partial \mathbf{F}_i}{\partial \mathbf{W}_Q} \approx a_{i i}\left(1-a_{i i}\right)\left[\frac{1}{\sqrt{d_k}} \mathbf{t}_i^{\top} \otimes\left(\mathbf{v}_i \mathbf{W}_V\right) \mathbf{W}_K^{\top} \mathbf{v}_i\right].
\end{equation}
Combining components:
\begin{equation}
\begin{aligned}
\frac{\partial \mathcal{L}_{\text {align }}}{\partial \mathbf{W}_Q} & \propto \sum_{i=1}^B \mathbf{v}_i^{\top}\left[a_{i i}\left(1-a_{i i}\right) \cdot \frac{1}{\sqrt{d_k}} \mathbf{t}_i^{\top} \otimes\left(\mathbf{v}_i \mathbf{W}_V\right) \mathbf{W}_K^{\top} \mathbf{v}_i\right] \mathbf{\Lambda}_i \\
& \approx \sum_{i=1}^B \mathbf{v}_i \mathbf{v}_i^{\top} \mathbf{t}_i \mathbf{t}_i^{\top} \mathbf{\Lambda}_i \cdot \underbrace{a_{i i}\left(1-a_{i i}\right) \frac{1}{\sqrt{d_k}}\left(\mathbf{W}_V \mathbf{W}_K^{\top} \mathbf{v}_i\right)}_{\text {scalar factor }}.
\end{aligned}
\end{equation}
Furthermore, the dominant term is:
\begin{equation}
\sum_{i=1}^B \mathbf{v}_i \mathbf{v}_i^{\top} \mathbf{t}_i \mathbf{t}_i^{\top} \mathbf{\Lambda}_i.
\end{equation}

\subsection{Dataset}
\begin{table*}[!t]
    \centering
    \begin{tabular}{c|ccc}
    \toprule
         Dataset&  \# Samples&  \# Hand-crafted features&  \# Clusters\\
         \midrule
         Fruit & 105& shape descriptors; color histogram& 3;3\\
         Fruit360 & 4,856& shape descriptors; color histogram& 4;4\\
         Card & 8,029& symbol shapes; color distribution& 13;4\\
         CMUface & 640 & HOG; edge maps& 4;20;2;4\\
         Standford Cars & 1,200& wheelbase length; body shape; color histogram& 4;3\\
         Flowers & 1,600& petal shape; color histogram& 4;4\\
         CIFAR-10 & 60,000& edge detection; color histograms; shape descriptors& 2;3\\
         \bottomrule
    \end{tabular}
     % \vskip -0.1in
    \caption{Statistics of the experimental datasets.}
    \label{tab:dataset_statistics}
\end{table*}
To demonstrate the effectiveness of Multi-DProxy, we conduct extensive evaluations across a diverse array of publicly available visual datasets commonly adopted for multi-clustering benchmarks \cite{yao2024multi}. This comprehensive datasets includes: \textbf{Stanford Cars} \cite{yao2024multi} (1,200 samples; clustering by color and vehicle type), \textbf{Card} \cite{yao2023augdmc} (8,029 samples; clustering by rank and suit), \textbf{CMUface} \cite{gunnemann2014smvc} (640 samples; clustering by pose, identity, glasses presence, and emotion), \textbf{Flowers} \cite{yao2024multi} (1,600 samples; clustering by color and species), \textbf{Fruit} \cite{hu2017finding} (105 samples; clustering by species and color), \textbf{Fruit360} \cite{yao2023augdmc} (4,856 samples; clustering by species and color), and \textbf{CIFAR-10} \cite{yao24customized} (clustering by object type and environment). These datasets represent standard evaluations capturing varied multi-clustering challenges. Detailed statistical information regarding data size, feature representations, and cluster configurations is summarized in Table~\ref{tab:dataset_statistics}.

Notably, some data may encounter challenges in extracting meaningful candidate categories from LLMs, or their labels may lack semantic features. For instance, in identity clustering on the CMUface dataset \cite{gunnemann2014smvc}, different identities represent distinct individuals, and the semantic meaning of names should not influence clustering outcomes. In such cases, following the widely used settings in previous works \cite{yao2024multi,yao24customized}, we randomly select candidate words from WordNet \cite{poli2010theory} as reference categories.

\subsection{Baselines}
We compare our proposed Multi-DProxy approach with eight state-of-the-art multiple clustering works. These works are as follows:
\begin{itemize}
    \item \textbf{MSC} \cite{hu2017finding}, which utilizes hand-crafted features to automatically identify distinct feature subspaces for different clustering.
    \item \textbf{MCV} \cite{guerin2018improving}, which employs multiple pre-trained feature extractors to represent different views of the same data.
    \item \textbf{ENRC} \cite{miklautz2020deep}, which integrates autoencoders with clustering objectives to generate diverse clustering.
    \item \textbf{iMClusts} \cite{ren2022diversified}, which leverages the representational power of deep autoencoders and multi-head attention to produce multiple salient embedding matrices and corresponding clustering.
    \item \textbf{AugDMC} \cite{yao2023augdmc}, which uses data augmentations to automatically extract features corresponding to various aspects of the data through a self-supervised prototype-based representation learning approach.
    \item \textbf{DDMC} \cite{yao2024dual}, which combines disentangled representation learning with a variational Expectation-Maximization (EM) framework.
    \item \textbf{Multi-DProxy} \cite{yao2024multi}, which relies on contrastive user-defined concepts to learn proxies tailored to user-specific interests.
    \item \textbf{Multi-Sub} \cite{yao24customized}, which incorporates multi-modal subspace proxy learning and leverages the synergistic capabilities of CLIP and GPT-4 to better capture user preferences.
\end{itemize}

In our experiments, we include both traditional and deep learning-based baselines. Traditional methods rely on hand-crafted features, whereas deep learning methods directly process the original images as input.

\end{document}